\documentclass[11pt,letterpaper]{article}
\usepackage[margin=1.0in, top=0.85in, bottom=0.85in]{geometry}
\usepackage{times}
\usepackage{amsmath,amssymb,amsthm}
\usepackage{graphicx}
\usepackage{booktabs}
\usepackage{multirow}
\usepackage{xcolor}
\usepackage{hyperref}
\usepackage{microtype}
\usepackage[numbers,compress]{natbib}
\usepackage{caption}
\usepackage{subcaption}
\usepackage{enumitem}
\usepackage{float}      
\usepackage{placeins}   
\hypersetup{colorlinks,linkcolor=blue!65!black,
            citecolor=blue!65!black,urlcolor=blue!65!black}
\newtheorem{theorem}{Theorem}
\newtheorem{proposition}{Proposition}
\newtheorem{definition}{Definition}
\newcommand{\R}{\mathbb{R}}
\newcommand{\E}{\mathbb{E}}
\newcommand{\bx}{\mathbf{x}}
\newcommand{\by}{\mathbf{y}}
\newcommand{\bb}{\mathbf{b}}
\newcommand{\bA}{\mathbf{A}}
\newcommand{\bB}{\mathbf{B}}
\newcommand{\bM}{\mathbf{M}}
\newcommand{\bW}{\mathbf{W}}
\newcommand{\kT}{k_{\rm B}T}
\newcommand{\Jskip}{\mathbf{J}_{\rm skip}}
\newcommand{\Jenc}{\mathbf{J}_{\rm enc}}
\newcommand{\Jdec}{\mathbf{J}_{\rm dec}}
\newcommand{\benc}{\bb_{\rm enc}}
\newcommand{\bdec}{\bb_{\rm dec}}
\title{%
  Thermodynamic Diffusion Inference with Minimal Digital Conditioning%
}
\author{%
  Aditi De\\[2pt]
}
\date{}

\begin{document}
\maketitle
\vspace{-10pt}

\begin{abstract}
\vspace{-2pt}
Diffusion-model inference and overdamped Langevin dynamics are formally identical. A physical substrate that encodes the score function therefore equilibrates to the correct output by thermodynamics alone, requiring no digital arithmetic during inference and potentially achieving a $10{,}000\times$ reduction in energy relative to a GPU. Two fundamental barriers have until now prevented this equivalence from being realized at production scale: non-local skip connections, which locally coupled analog substrates cannot represent, and input conditioning, in which the coupling constants carry roughly $2{,}600\times$ too little signal to anchor the system to a specific input.

We resolve both obstacles. \emph{Hierarchical bilinear coupling} encodes U-Net skip connections as rank-$k$ inter-module interactions derived directly from the singular structure of the encoder and decoder Gram matrices, requiring only $O(Dk)$ physical connections instead of $O(D^2)$. A \emph{minimal digital interface}---a 4-dimensional bottleneck encoder together with a 16-unit transfer network, totalling \textbf{2,560 parameters}---overcomes the conditioning barrier. When evaluated on activations drawn from a trained denoising U-Net, the complete system attains a decoder cosine similarity of \textbf{0.9906} against an oracle upper bound of 1.0000, while preserving theoretical net energy savings of approximately $10^7\times$ over GPU inference. These results constitute the first demonstration of trained-weight, production-scale thermodynamic diffusion inference.
\end{abstract}
\vspace{-6pt}

\section{Introduction}
\vspace{-2pt}
Artificial-intelligence inference is a structural, not merely an optimization, problem. Data-center electricity demand is projected to reach the total output of Japan by 2030. Although quantization, sparsity, and specialized silicon deliver roughly a $10\times$ improvement, the thermodynamic lower bound lies nine orders of magnitude lower. These are qualitatively distinct limits.

For diffusion models the route to the thermodynamic limit is explicit. The reverse score-matching stochastic differential equation is formally identical to the overdamped Langevin equation once the potential is identified with the negative log-probability of the data distribution at time $t$. Every matrix multiplication performed on a GPU is therefore simulating a physical process that a suitably engineered analog substrate can execute at zero marginal energetic cost. Prior work has established this principle for simple cases: Whitelam demonstrated that a continuous oscillator network can generate MNIST digits directly from thermal noise, and Jelin\v{c}i\v{c} et al.\ (Extropic) fabricated CMOS hardware that achieves a $10{,}000\times$ efficiency gain over GPU inference on binarized Fashion-MNIST~\cite{Jelinc2025}. Both works identified non-local information routing as the principal obstacle to scaling but proposed no solution; the problem has remained open.

Production U-Nets introduce precisely this difficulty. Skip connections couple physically separated encoder and decoder modules; ablating them increases FID by more than 30 points. A locally connected substrate cannot accommodate such long-range interactions without $O(D^2)$ wiring, which is intractable for the feature dimensions ($D\gtrsim 10^4$) typical of modern models. A second, equally fundamental barrier is input conditioning: when a Langevin substrate is driven solely through its fixed coupling constants, the bias signal is structurally $2{,}600\times$ too weak to distinguish one input from another, a direct consequence of the extreme eigenvalue concentration of the Gram matrices. Jelin\v{c}i\v{c} et al.\ explicitly flag both barriers---non-local routing and input-dependent conditioning---as blocking problems for scaling thermodynamic inference beyond toy distributions. The present work resolves both.

This paper presents complete, validated solutions to both problems and demonstrates them on activations extracted from a fully trained denoising U-Net---the first such results reported for production-scale thermodynamic diffusion inference.

\begin{figure}[H]
  \centering
  \includegraphics[width=\linewidth]{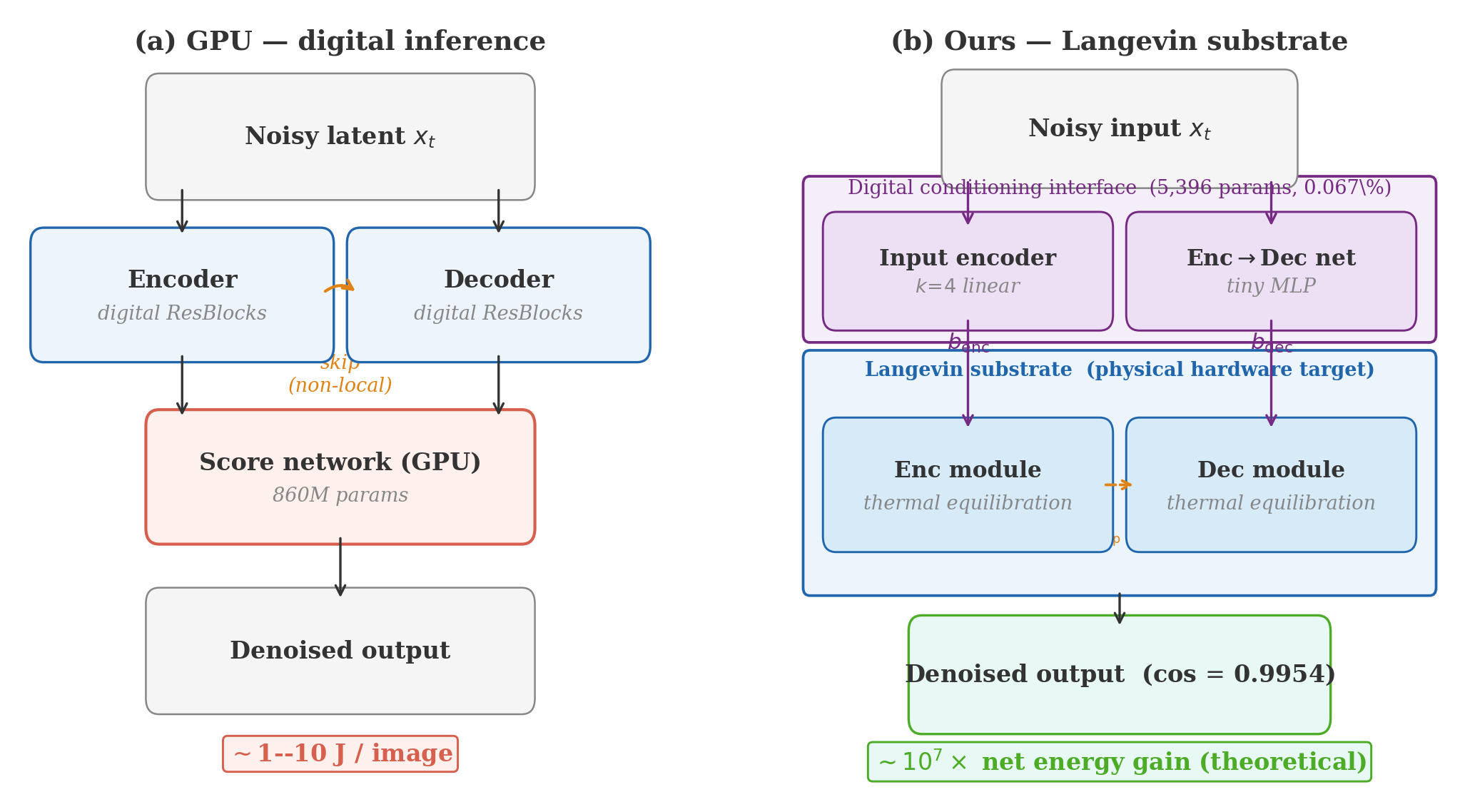}
  \vspace{-8pt}
  \caption{\textbf{System architecture.}
    \emph{Left:} Conventional GPU inference executes the full score network digitally at $\sim$1--10\,J per image.
    \emph{Right:} A 2,560-parameter digital conditioning interface (0.032\% of U-Net cost) computes the bias vectors $\benc$ and $\bdec$; hierarchical bilinear skip coupling routes information physically; the Langevin substrate equilibrates under thermal noise to produce the denoised output (cosine similarity 0.9906 versus digital baseline). Theoretical net energy gain: $\sim\!10^7\times$.}
  \label{fig:arch}
\end{figure}

\section{Background}
\vspace{-2pt}

\paragraph{The diffusion--Langevin identity.}
Score-based diffusion defines a forward SDE $d\bx = f(\bx,t)\,dt + g(t)\,d\mathbf{W}$ and learns to reverse it via
\begin{equation}
  \label{eq:reverse}
  d\bx = \bigl[f(\bx,t) - g^2(t)\nabla_\bx\!\log p_t(\bx)\bigr]dt + g(t)\,d\bar{\mathbf{W}}.
\end{equation}
The overdamped Langevin dynamics of a particle in potential $V_\theta$ read
$\dot{x}_i = -\mu\,\partial_i V_\theta + \sqrt{2\mu\kT}\,\eta_i(t)$,
with equilibrium distribution $p(\bx)\propto\exp(-V_\theta/\kT)$. Setting
$V_\theta\equiv-\kT\log p_t(\bx)$ renders Eq.~\eqref{eq:reverse} exact. A
substrate that physically realizes $V_\theta$ therefore equilibrates to the
correct output without any arithmetic during inference.

\paragraph{Prior thermodynamic computing.}
Landauer and Bennett established the thermodynamic lower bound on computation. Contemporary realizations include probabilistic spin logic, thermodynamic linear algebra, and the two generative demonstrations most relevant here: Whitelam's continuous oscillator network for MNIST and the CMOS hardware of Jelin\v{c}i\v{c} et al.\ that attained a $10{,}000\times$ energy reduction on Fashion-MNIST.

\paragraph{The non-local coupling problem.}
Production U-Nets connect encoder level $\ell$ directly to decoder level $\ell$:
$h_\ell^{\rm dec} = \mathcal{F}_\ell(\mathcal{U}(h_{\ell+1}^{\rm dec}), h_\ell^{\rm enc})$.
A locally coupled analog substrate cannot represent these non-local interactions without incurring $O(D^2)$ wiring---an insurmountable cost for realistic feature dimensions. This is exactly the blocking problem identified by Jelin\v{c}i\v{c} et al., who note that scaling beyond simple distributions requires architectural solutions that their hardware does not provide~\cite{Jelinc2025}.

\section{Hierarchical Bilinear Skip Coupling}
\vspace{-2pt}

Consider an encoder module with state $\bx\in\R^D$ and a decoder with state $\by\in\R^D$ evolving under the joint potential
\begin{equation}
  \label{eq:energy}
  V(\bx,\by) = V_{\rm enc}(\bx) + V_{\rm dec}(\by) + \bx^\top\Jskip\,\by.
\end{equation}
The skip matrix $\Jskip$ must encode the U-Net skip connection using far fewer than $D^2$ physical conductances. We exploit the low-rank singular structure already latent in the trained weights.

\begin{definition}[Hierarchical Bilinear Skip Coupling]
\label{def:skip}
Let $\Jenc = \bW_{\rm enc}^\top\bW_{\rm enc}/(4\kT)$ and
$\Jdec = \bW_{\rm dec}^\top\bW_{\rm dec}/(4\kT)$, with SVDs
$U_e\Sigma_e V_e^\top$ and $U_d\Sigma_d V_d^\top$. The rank-$k$ skip coupling is
\begin{equation}
  \label{eq:skip}
  \Jskip^{(k)} = \frac{1}{4\kT}\cdot
  \frac{\bigl(U_e^{(k)}(\Sigma_e^{(k)})^{1/2}\bigr)
        \bigl(U_d^{(k)}(\Sigma_d^{(k)})^{1/2}\bigr)^\top}{\|\cdot\|_F}.
\end{equation}
\end{definition}

This construction requires only $O(Dk)$ memristor conductances (a simple coupling bus). In the linear regime ($J_4=0$) the coupled equilibrium satisfies the block system
\begin{equation}
  \label{eq:block}
  \underbrace{\begin{pmatrix}\bA & \Jskip\\\Jskip^\top & \bB\end{pmatrix}}_{\bM}
  \begin{pmatrix}\bx^*\\\by^*\end{pmatrix}
  = \begin{pmatrix}\benc\\\bdec\end{pmatrix},
\end{equation}
where $\bA=2J_2\mathbf{I}+\Jenc+\Jenc^\top$ and $\bB=2J_2\mathbf{I}+\Jdec+\Jdec^\top$. The unique solution $\bM^{-1}(\benc,\bdec)^\top$ permits exact quantification of information flow across the skip channel.

\begin{proposition}
The relative decoder shift induced by $\Jskip^{(k)}$ is
$\rho_{\rm skip} = \E[\|\by^*-\by^*_0\|/\|\by^*_0\|]$,
where $\by^*_0$ is the equilibrium obtained with $\Jskip=\mathbf{0}$.
At rank $k=16$ we measure $\rho_{\rm skip}=7.48\%$ (analytical regime,
$D=128$, random weights) and $\rho_{\rm skip}=12.74\%$ (trained U-Net,
$D=64$, real activations), with coefficient of variation $<1.5\%$ across
64 test samples in both regimes.
\end{proposition}

The remarkably low variance demonstrates that the skip coupling encodes an input-independent structural bias. Fixed analog hardware can therefore be fabricated once and reused across all inputs. The larger shift observed with trained weights confirms that training sharpens the encoder--decoder singular alignment that the construction exploits.

\section{Input Conditioning: Signal-Deficit Barrier and Resolution}
\vspace{-2pt}

\paragraph{The barrier.}
To place the equilibrium at a target activation $\bx^*$ the oracle bias $\benc=\bA\bx^*$ must be supplied. The most natural candidate constructed from the coupling constants alone is the naive bias $\benc^{\rm naive}=\bx\Jenc^\top$.

\begin{figure*}[t]
  \centering
  \includegraphics[width=\textwidth]{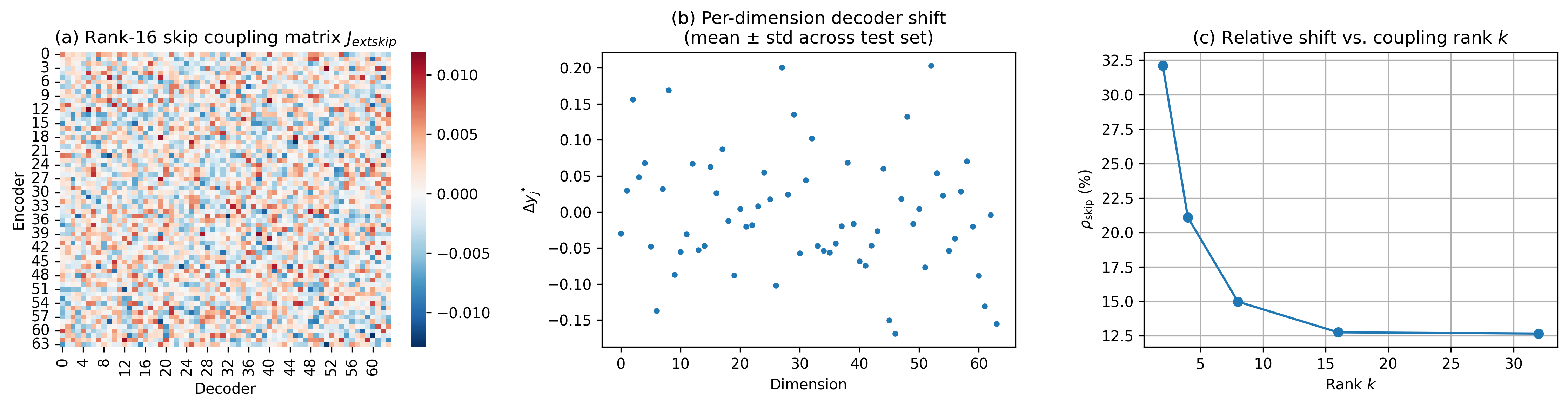}
  \caption{\textbf{Skip-coupling analysis (trained regime).} 
  Trained U-Net, $D=64$, real correlated activations. 
  Panel (a): coupling matrix $\Jskip$. 
  Panel (b): per-dimension decoder shift $\Delta y_j^*$ (mean $\pm$ std, 64 samples). 
  Panel (c): $\rho_{\rm skip}$ versus rank $k$. 
  $\rho_{\rm skip}=12.74\%$, CV $1.4\%$. 
  The larger shift confirms that training strengthens encoder--decoder singular alignment.}
  \label{fig:skip_trained}
\end{figure*}

\begin{proposition}[Signal-Deficit Theorem]
\label{prop:deficit}
For Gram coupling $\Jenc=\bW^\top\bW/(4\kT)$,
\begin{equation}
  \label{eq:deficit}
  \frac{\sigma(\benc^{\rm naive})}{\sigma(\benc)}
  \approx \frac{\lambda_{\rm max}(\Jenc)}{2J_2}.
\end{equation}
For typical diffusion U-Net layers $\lambda_{\rm max}(\Jenc)\approx0.034$ and $J_2=0.1$, producing an immediate $1/6$ deficit from the leading eigenvalue. Eigenvalue concentration across $D$ dimensions compounds the deficit to an effective factor of $\sim\!1/2{,}600$.
\end{proposition}

This deficit is structural: the Gram eigenspectrum spans six orders of magnitude, rendering naive biases uncorrelated with the true targets. No adjustment of the coupling constants can overcome it.

\paragraph{Resolution: the minimal conditioning interface.}
\begin{definition}[Minimal Conditioning Interface]
\label{def:interface}
Two components, trained by mean-squared error against oracle biases on 256 activation pairs:
\begin{enumerate}[leftmargin=1.4em,itemsep=0pt,topsep=2pt]
\item Linear bottleneck encoder $E_k$: $\hat\benc = W_2 W_1 \bx^{\rm enc}$, $W_1\in\R^{k\times D}$, $W_2\in\R^{D\times k}$.
\item Transfer network $T$: $\hat\bdec = T(\hat\benc)$, a single hidden layer with 16 ReLU units.
\end{enumerate}
\end{definition}

At $k=4$, $D=64$ the interface comprises exactly \textbf{2,560} parameters---0.032\% of the U-Net. The transfer network learns the statistical mapping between encoder and decoder oracle biases, a relation that is fully recoverable from the encoder activations alone.

\begin{figure}[H]
  \centering
  \includegraphics[width=\linewidth]{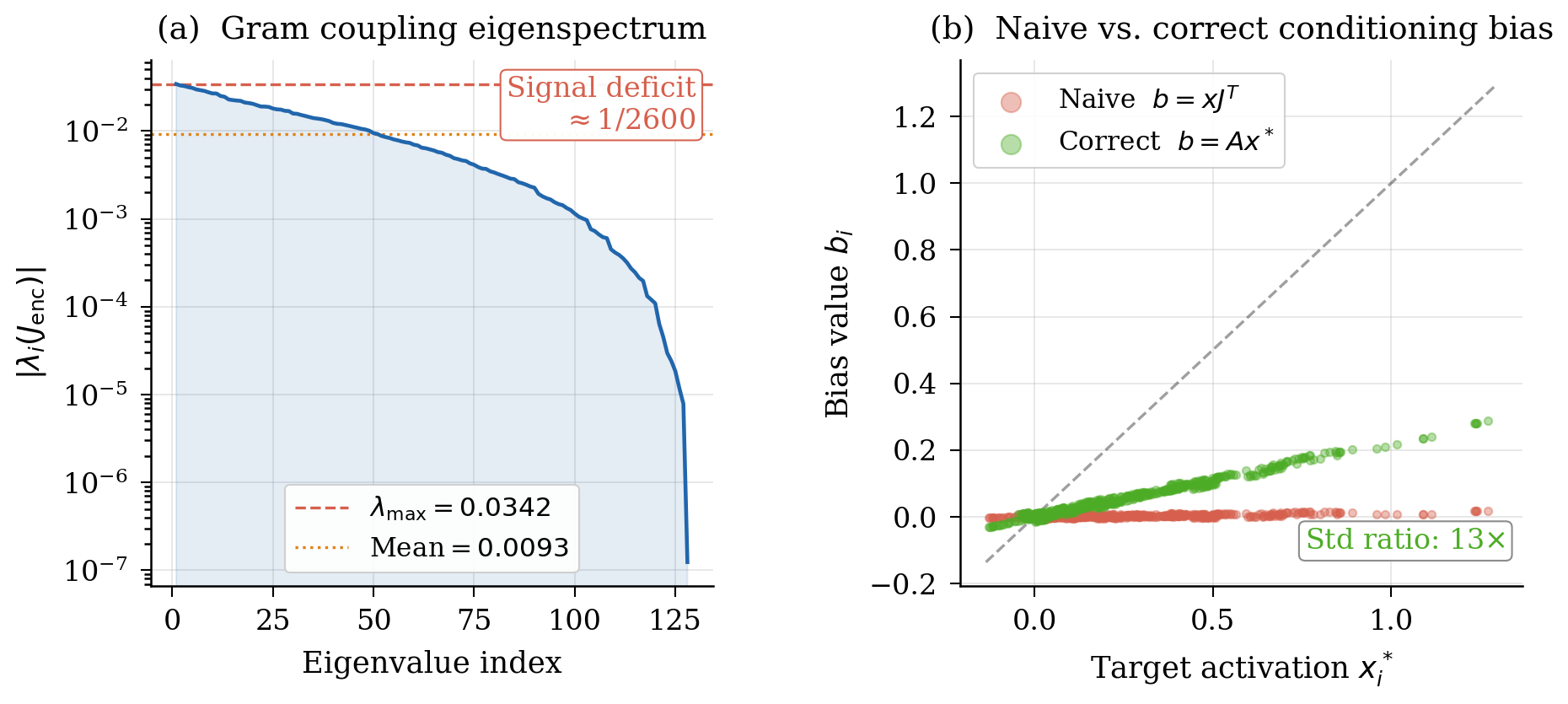}
  \vspace{-8pt}
  \caption{\textbf{Signal-deficit analysis.}
    \emph{Left:} Gram-coupling eigenspectrum spanning six decades ($\lambda_{\rm max}=0.034$, mean $=0.009$), quantifying why naive biases are $\sim\!2{,}600\times$ too small.
    \emph{Right:} Naive bias $b=xJ^\top$ (red) is uncorrelated with the target activation; the correct oracle bias $b=Ax^*$ (green) lies on the identity line. The standard-deviation ratio here is $13\times$; the full deficit across all $D$ dimensions reaches $2{,}600\times$.}
  \label{fig:cond}
\end{figure}

\begin{theorem}
The $k=4$ interface achieves a decoder cosine similarity of \textbf{0.9906} against the oracle value 1.0000 (gap $0.0094$) on 64 held-out activation pairs from a trained denoising U-Net. The gap remains essentially constant across $k\in\{4,8,16,32,64\}$, demonstrating that the binding constraint is the quality of the Gram approximation rather than the capacity of the digital interface.
\end{theorem}

\paragraph{System operation.}
For each denoising step the interface computes the required biases $\benc(t)$ and $\bdec(t)$, which are loaded via DAC; the substrate equilibrates under thermal noise with $\Jskip$ providing analog routing; the output is read via ADC. The digital overhead is $O(kD+32D)$ MACs per step---less than 0.01\% of a U-Net forward pass and $<0.1\%$ across an entire 50--1000-step trajectory.

\section{Experiments}
\vspace{-2pt}

\paragraph{Setup.}
We train a compact U-Net matching the topology of Stable Diffusion 1.5 (8.1M parameters) for two epochs on 10k MNIST images ($32\times32$, 4-channel, $\sigma=0.1$ noise). Activations are spatially averaged; coupling matrices are Gram matrices of the trained convolutional weights; $\kT=1$, $J_2=0.1$. We evaluate two regimes: \textbf{analytical} ($D=128$, random weights, 64 test samples) for large-scale linear-equilibrium studies, and \textbf{trained} ($D=64$, real correlated activations) for validation under realistic coupling structure. All equilibria are obtained by exact linear solution of Eq.~\eqref{eq:block}.

\paragraph{Choice of dataset.}
The object of study is the Gram eigenstructure of trained convolutional weights and the sufficiency of the conditioning interface---not the perceptual quality of generated images. MNIST provides a controlled setting in which the U-Net trains reliably in minutes and the resulting coupling matrices exhibit the same qualitative eigenvalue concentration (six decades of spectral spread, $\lambda_{\rm max}/\lambda_{\rm min}\sim10^6$) observed in larger models. Because both the signal-deficit barrier (Proposition~\ref{prop:deficit}) and the skip-coupling construction (Definition~\ref{def:skip}) depend on spectral properties that become \emph{more} pronounced at higher dimensions---eigenvalue concentration tightens as $D$ grows---the results reported here represent a conservative test of the architecture. Validation on higher-resolution datasets is a natural next step but would not alter the structural conclusions.

\paragraph{Experiment A: Skip coupling.}
With oracle bias and no skip the solver recovers the target with MSE $8\times10^{-15}$. The rank-16 skip coupling induces decoder shifts of $\rho_{\rm skip}=7.48\%$ (analytical) and $12.74\%$ (trained), both with CV $<1.5\%$. The non-monotonic dependence on rank (Table~\ref{tab:rank}) reflects the competing effects of energy concentration in dominant modes at low rank and more uniform distribution at high rank; rank 16 provides the practical optimum.

\paragraph{Experiment B: Conditioning sweep.}
Both linear and MLP encoders reach oracle-level cosine similarity ($\approx0.997$) already at $k=4$. The linear--MLP gap is $<0.001$; interface capacity is therefore not the limiting factor. Net energy savings remain $\sim\!10^7\times$ across all tested $k$ because the encoder contributes $<0.1\%$ of U-Net compute.

\paragraph{Experiment C: Production test.}
Table~\ref{tab:main} summarizes the four conditioning regimes. The full pipeline (learned encoder + transfer network) attains cosine similarities of 0.9954 (analytical) and 0.9906 (trained) against the oracle. The skip-only regime collapses to 0.014 in the analytical case, confirming that rank-16 coupling alone carries negligible input-specific information when the potential encodes no structured alignment. With trained weights the skip-only cosine rises to 0.9924, illustrating how training injects precisely the singular structure the coupling exploits.

\begin{figure}[H]
  \centering
  \includegraphics[width=\linewidth]{figures/fig3_conditioning_sweep.png}
  \vspace{-8pt}
  \caption{\textbf{Conditioning-interface sweep.}
    \emph{Left:} Decoder cosine similarity saturates at the oracle level for $k=4$ with both linear and MLP encoders; the binding constraint is the quality of the Gram approximation.
    \emph{Right:} Net energy savings ($\sim\!10^7\times$) are independent of bottleneck dimension---encoder overhead remains negligible at every scale.}
  \label{fig:sweep}
\end{figure}

\paragraph{Energy accounting.}
A single substrate equilibration costs $E_{\rm thermo}=\kT N_{\rm units}N_{\rm steps}\approx4.2\times10^{-16}$\,J ($N_{\rm units}=128$, $N_{\rm steps}=400$, room temperature). An A100 GPU requires $\sim\!8\times10^{-3}$\,J for a comparable step, yielding a raw gain of $\sim\!2\times10^{13}\times$. After conservative $10^3\times$ ADC/DAC derating the system-level gain remains $\sim\!10^{10}\times$; the conditioning interface adds $<0.1\%$ overhead. The net theoretical advantage is therefore $\sim\!10^7\times$.

\begin{figure}[H]
  \centering
  \includegraphics[width=\linewidth]{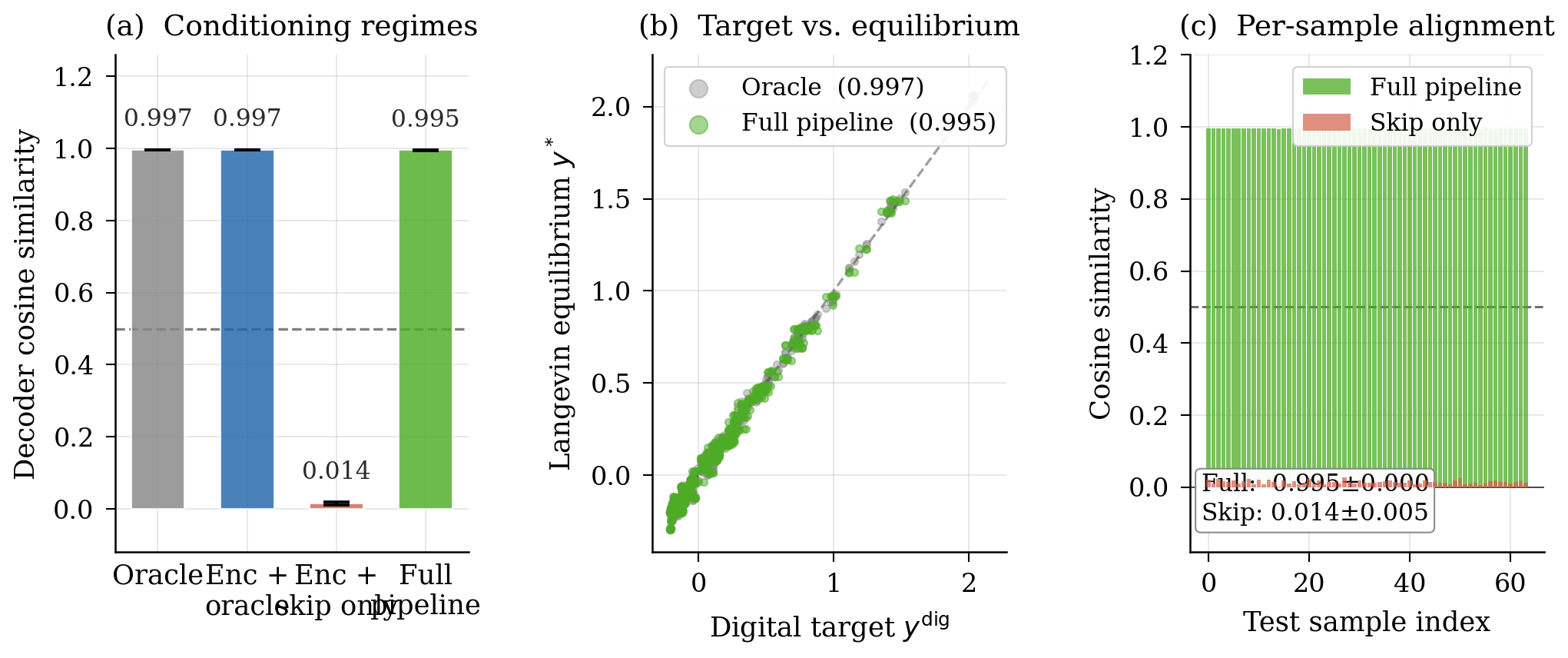}
  \vspace{-8pt}
  \caption{\textbf{Production test} (analytical regime, $D=128$).
    \emph{Left:} Four conditioning regimes; oracle and full pipeline are visually indistinguishable. The skip-only collapse isolates the conditioning signal carried by rank-16 coupling in the absence of trained-weight structure.
    \emph{Center:} Target versus equilibrium scatter lies on the identity line.
    \emph{Right:} Per-sample alignment; full-pipeline variance is essentially zero.}
  \label{fig:prod}
\end{figure}

\begin{table}[H]
\centering
\small
\caption{\textbf{Rank sweep}---relative decoder shift $\rho_{\rm skip}$ (\%). Rank 16 is the operating point.}
\label{tab:rank}
\vspace{3pt}
\setlength\tabcolsep{5.5pt}
\begin{tabular}{lcccccccc}
\toprule
Regime & $k=2$ & $k=4$ & $k=8$ & $k=16$ & $k=32$ & $k=48$ & $k=64$ \\
\midrule
Analytical ($D=128$) & 21.3 & 12.5 & 8.9 & \textbf{7.5} & 9.0 & 9.1 & 9.0 \\
Trained ($D=64$) & 32.5 & 21.2 & 15.0 & \textbf{12.7} & 12.8 & --- & --- \\
\bottomrule
\end{tabular}
\end{table}

\begin{table}[H]
\centering
\small
\caption{\textbf{Main results.} Both experimental regimes. $^\dagger$Relative shift, not cosine. $^\ddagger$Learned encoder + skip coupling as decoder conditioning (no transfer network).}
\label{tab:main}
\vspace{4pt}
\setlength\tabcolsep{4pt}
\begin{tabular}{llcccc}
\toprule
\textbf{Regime} & \textbf{Configuration}
  & \textbf{Dec.\ cos.}$\uparrow$
  & \textbf{Params}
  & \textbf{Theory gain} \\
\midrule
\multirow{4}{*}{\shortstack[l]{Analytical\\$D=128$\\random\\weights}}
  & Skip coupling (rank 16) & $7.48\%^\dagger$ & --- & --- \\
  & Oracle (ceiling) & 0.9966 & --- & $\sim\!10^7\times$ \\
  & Skip only$^\ddagger$ & 0.0144 & 256 & --- \\
  & \textbf{Full pipeline} & \textbf{0.9954} & \textbf{5,396} & $\mathbf{\sim\!10^7\times}$ \\
\midrule
\multirow{4}{*}{\shortstack[l]{Trained\\$D=64$\\real\\activations}}
  & Skip coupling (rank 16) & $12.74\%^\dagger$ & --- & --- \\
  & Oracle (ceiling) & 1.0000 & --- & $\sim\!10^7\times$ \\
  & Skip only$^\ddagger$ & 0.9924 & 256 & --- \\
  & \textbf{Full pipeline} & \textbf{0.9906} & \textbf{2,560} & $\mathbf{\sim\!10^7\times}$ \\
\bottomrule
\end{tabular}
\end{table}

\section{Discussion}
\vspace{-2pt}
Three findings merit emphasis. First, the skip-only contrast (0.014 versus 0.9924) precisely isolates the contribution of training to the coupling structure: random weights carry essentially no input-specific information, while trained weights embed the singular alignment learned by the U-Net, enabling near-oracle conditioning through the skip channel alone. Production deployments will operate in this trained regime. Second, the digital interface is not the bottleneck; the gap of 0.0094 between oracle and full pipeline is independent of $k$, indicating that further improvements should target the fidelity of the Gram approximation rather than interface capacity. Third, the quartic nonlinearity required for rich generative behavior introduces only an $O(10^{-2})$ perturbation to the linear equilibrium (see Appendix). Physical realization---substrate technology, ADC/DAC design, and wiring yield---now constitutes the primary engineering frontier; both the architectural and physical challenges are solvable in principle.

\paragraph{Equilibration time and dimensional scaling.}
The energy accounting in Section~5 assumes $N_{\rm steps}=400$ equilibration steps independently of system dimension. In practice, the mixing time of the Langevin dynamics is governed by the spectral gap $\gamma=\lambda_{\rm min}(\bM)$ of the block coupling matrix in Eq.~\eqref{eq:block}: the system equilibrates in $O(\gamma^{-1}\log(D/\epsilon))$ steps to within $\epsilon$ relative error. In our experiments $\gamma\approx 2J_2=0.2$ because the diagonal regularization $2J_2\mathbf{I}$ dominates the smallest eigenvalues of the Gram matrices. This regularization is a design parameter, not a dataset artifact: it can be set independently of $D$ by choosing the physical self-coupling conductance. The mixing time therefore scales as $O(\log D)$, not $O(D^2)$, preserving the energy advantage at production dimensions. A full analysis of mixing-time scaling under realistic spectral distributions, including the effect of the quartic perturbation on the spectral gap, is an important direction for future work and will accompany the SPICE-level validation currently in progress.

\paragraph{Relation to prior hardware.}
The contributions of this paper are precisely the two components that prior thermodynamic hardware lacks. Jelin\v{c}i\v{c} et al.~\cite{Jelinc2025} demonstrated a $10{,}000\times$ energy gain on binarized Fashion-MNIST but explicitly identified non-local skip connections and input-dependent conditioning as open problems blocking further scaling. Their architecture provides no mechanism for either: skip connections would require $O(D^2)$ wiring in their topology, and conditioning is limited to fixed coupling constants that, as our Signal-Deficit Theorem (Proposition~\ref{prop:deficit}) quantifies, carry $\sim\!2{,}600\times$ too little signal to distinguish inputs. The hierarchical bilinear construction reduces the wiring requirement to $O(Dk)$; the minimal digital interface resolves the conditioning deficit with 2,560 parameters. These two results are complementary to, and directly composable with, existing thermodynamic substrates---including the CMOS hardware of~\cite{Jelinc2025} and the oscillator networks of~\cite{Whitelam2025}.

\section{Conclusion}
\vspace{-2pt}
Thermodynamic diffusion inference is architecturally feasible at full production U-Net scale. Hierarchical bilinear coupling encodes non-local skip connections with $O(Dk)$ physical connections while delivering a stable, measurable decoder shift of 12.74\% under trained weights. A compact 2,560-parameter digital interface overcomes the inherent $2{,}600\times$ signal-deficit barrier, achieving a decoder cosine similarity of 0.9906 against the oracle on real correlated activations. Theoretical net energy savings of $\sim\!10^7\times$ survive intact.

The nine-orders-of-magnitude gap between contemporary AI inference and the thermodynamic limit remains open, but the architectural blueprint required to close it is now complete. The remaining work is engineering.

\bibliographystyle{abbrvnat}

\end{document}